%% file: main.tex
\documentclass[10pt,twocolumn,letterpaper]{article}

\usepackage[pagenumbers]{cvpr} 


\definecolor{cvprblue}{rgb}{0.21,0.49,0.74}
\definecolor{bg-blue}{HTML}{ebe6ef}
\definecolor{lightgrey}{rgb}{0.43,0.43,0.43}
\usepackage[pagebackref,breaklinks,colorlinks,allcolors=cvprblue]{hyperref}
\usepackage{xspace}
\usepackage{amsmath,amsfonts,amssymb}
\usepackage{subcaption}
\usepackage{booktabs}
\usepackage{caption}
\usepackage{adjustbox}
\usepackage{xcolor}
\usepackage{colortbl}
\usepackage{pifont}
\usepackage{bbding}
\usepackage{multirow}
\usepackage{sidecap}

\newcommand{\cmark}{\ding{51}}%
\newcommand{\xmark}{\ding{55}}%

\def\paperID{13505} 
\def\confName{CVPR}
\def\confYear{2025}

\title{Token Cropr: Faster ViTs for Quite a Few Tasks}

\author{Benjamin Bergner\textsuperscript{\textnormal{1}}, Christoph Lippert\textsuperscript{\textnormal{1,2}}, Aravindh Mahendran\textsuperscript{\textnormal{3}} \\
\textsuperscript{1}Hasso Plattner Institute for Digital Engineering, University of Potsdam\\
\textsuperscript{2}Hasso Plattner Institute for Digital Health at the Icahn School of Medicine at Mount Sinai\\
\textsuperscript{3}Google DeepMind
}

\newcommand{\methodname}{Cropr\xspace}
\newcommand{\R}{\mathbb{R}}
\newcommand*{\belowrulesepcolor}[1]{%
  \noalign{%
    \kern-\belowrulesep
    \begingroup
      \color{#1}%
      \hrule height\belowrulesep
    \endgroup
  }%
}
\newcommand*{\aboverulesepcolor}[1]{%
  \noalign{%
    \begingroup
      \color{#1}%
      \hrule height\aboverulesep
    \endgroup
    \kern-\aboverulesep
  }%
}
\newcommand{\stair}{\scalebox{1.2}{\rotatebox[origin=c]{270}{$\Rsh$}}}
\newcommand{\drarrow}{\scalebox{0.8}{$\searrow$}}
\newcommand{\ebar}[2]{$#1$ {\scriptsize $#2 \times$}} 

\begin{document}
\maketitle
\input{sec/0_abstract}    
\input{sec/1_Introduction}
\input{sec/2_Related_Work}
\input{sec/3_Methods}
\input{sec/4_Experiments}
\input{sec/5_Conclusion}
\clearpage
{
    \small
    \bibliographystyle{ieeenat_fullname}
    \bibliography{main}
}


\maketitlesupplementary
\input{sec/X_suppl}

\end{document}

%% file: sec/0_abstract.tex
\begin{abstract}
The adoption of Vision Transformers (ViTs) in resource-constrained applications necessitates improvements in inference throughput.
To this end several token pruning and merging approaches have been proposed that improve efficiency by successively reducing the number of tokens.
However, it remains an open problem to design a token reduction method that is fast, maintains high performance, and is applicable to various vision tasks.
In this work, we present a token pruner that uses auxiliary prediction heads that learn to select tokens end-to-end based on task relevance.
These auxiliary heads can be removed after training, leading to through\-put close to that of a random pruner.
We evaluate our method on image classification, semantic segmentation, object detection, and instance segmentation, and show speedups of 1.5 -- 4x with small drops in performance.
As a best case, on the ADE20k semantic segmentation benchmark, we observe a 2x speedup relative to the no-pruning baseline, with a negligible performance penalty of 0.1 median mIoU across 5 seeds.
\end{abstract}

%% file: sec/1_Introduction.tex
\section{Introduction}
\label{sec:intro}

The Vision Transformer~\cite{dosovitskiy2021an} is a widely used architecture for computer vision tasks such as image classification, segmentation, and object detection~\cite{vit_survey}.
ViTs represent images as a sequence of per-patch tokens, that they process using multi-head self-attention (MHSA) transformer blocks.
The self-attention mechanism computes a pairwise dot product between all tokens, which results in a quadratic time and space complexity with respect to sequence length, $\mathcal{O}\!\left(n^2\right)$~\cite{vaswani2017attention}.
For real-world applications that require low latency or a small compute budget, sequence length thus becomes a burden, especially in the light of increasing model sizes~\cite{pmlr-v202-dehghani23a}, image resolutions~\cite{bergner2022iterative}, and finer tokenization~\cite{beyer2023flexivit}.

\begin{figure}[t]
  \centering
  \includegraphics[width=0.85\linewidth]{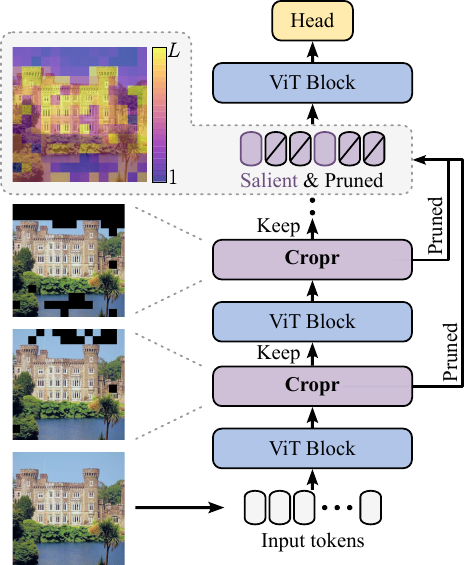}
   \caption{\textbf{Cro}ss-attention \textbf{pr}uning (\textbf{\methodname}) modules successively prune less relevant tokens, retaining only the most discriminative ones for deeper layers. Our method accelerates ViTs while maintaining high performance and is applicable to many vision tasks, from classification to segmentation and detection. The example castle images illustrate the pruning process. The heatmap visualizes which tokens were pruned at each block $1$ to $L$ in the network.}
   \label{fig:cropr_overview}
\end{figure}

Images, such as the example used in~\cref{fig:cropr_overview}, are spatially redundant containing non-salient background and repetitive patterns.
This suggests that several patches could be processed using fewer transformer blocks, providing an opportunity to prune uninformative tokens, reduce sequence length in higher layers, and thus improve computational efficiency.
However, this raises a central question: How can we accurately and efficiently assess the importance of individual tokens for a given task?

Recent token pruning methods rely on heuristics, such as self-attention scores, to identify informative tokens~\cite{liang2022not, fayyaz2022adaptive}.
Alternative approaches reduce token count by merging similar tokens~\cite{bolyatoken, marin2023token, du2016study}.
However, these methods do not explicitly model the importance of a token for a given task, which can lead to a significant drop in task performance.
In contrast, attribution methods such as Saliency~\cite{simonyan2013deep}, Occlusion~\cite{zeiler2014visualizing} and Attention Rollout~\cite{abnar-zuidema-2020-quantifying} estimate input contributions to a prediction, but require a full forward pass, which is not a viable option due to the associated overhead.

The question of estimating task relevance is further complicated by the diversity of task types in computer vision.
Image classification, the simplest of vision tasks, has been the focus of many prior works,~\cite{liang2022not, long2023beyond, xu2022evo, kong2022spvit, rao2021dynamicvit, fayyaz2022adaptive, yin2022vit, pan2021ia} to name a few.
Dense tasks such as semantic segmentation, however, present a new challenge for token pruning since they require predictions at the pixel level, which is inherently in conflict with the idea of pruning tokens.

We address these concerns with \textbf{Cro}ss-attention \textbf{pr}uning (\textbf{\methodname}), a simple token pruning method for ViTs that efficiently estimates per-token task relevance, while being applicable to various vision tasks.
\methodname modules are applied at intermediate layers for token pruning, see~\cref{fig:cropr_overview}.
Each module consists of a cross-attention based aggregation mechanism coupled with an auxiliary pred\-iction head.
The latter learns to solve the task while the former ranks tokens by task relevance, forwarding only the most relevant tokens to deeper layers.
The auxiliary heads can be discarded after training, which minimizes overhead and renders token pruning efficient.
Lastly, pruned tokens are reintroduced later in the network, in a trick called Last Layer Fusion, to enable dense tasks.
We detail our method in~\cref{sec:method}.

We evaluate our method on image classification, semantic segmentation, object detection, and instance segmentation in~\cref{sec:experiments}.
We demonstrate strong performance even under aggressive pruning schedules.
For example, when fine-tuning an EVA-02~\cite{eva02} backbone, we are able to maintain $89.7$\% top-1-accuracy on ImageNet-1k, a drop of only $0.2$ percentage points compared to the unpruned model, while achieving a $2.1\times$ speedup.
We also evaluate the effect of token pruning on different encoder capacities and image resolutions, showing that our method performs particularly well at scale.
An ablation study offers empirical support for our design choices and qualitative evaluations provide insights into the pruning process.
We conclude in~\cref{sec:conc} with a summary of our findings and future work.

%% file: sec/2_Related_Work.tex
\section{Related Work}
\label{sec:related}
Several methods have been proposed to reduce sequence length in vision transformers by pruning / merging tokens. 

\vspace{-0.51em}
\paragraph{Token pruning for classification.}
A common strategy is to leverage attention scores from the class token (CLS) into image tokens as a bottom-up cue for pruning~\cite{liang2022not, fayyaz2022adaptive, long2023beyond, xu2022evo}.
Tokens with lower attention scores are regarded as less important and pruned out.
Notably,~\citet{Haurum_2023_ICCV} show that a simple Top-K selector is a strong baseline.
However, modern fused kernel implementations~\cite{dao2022flashattention, rabe2021selfattention} often restrict direct access to attention matrices, thus requiring alternative strategies.
We instead take a top-down approach, leveraging signals from auxiliary heads to retain task-relevant tokens.

Other approaches use parametrized modules to predict which tokens to keep~\cite{kong2022spvit, rao2021dynamicvit}, 
but introduce additional layers and losses that may interfere with the primary task.
\methodname modules apply a stop-gradient to avoid gradient interference and limit additional parameters to a single query token at inference time.
Another common design choice is to make token pruning adaptive, pruning more tokens for simpler inputs~\cite{fayyaz2022adaptive, yin2022vit, pan2021ia, li2022sait, tang2022patch, dong2023heatvit, kim2022learned}.
In contrast, we use a throughput-optimized static approach that prunes a constant number of tokens to enable batching across inputs.

\vspace{-0.51em}
\paragraph{Token pruning beyond classification.} Very few works apply token pruning beyond classification:
\citet{tang2023dynamic, liu2024dynamic}~extend it to semantic segmentation by adding auxiliary heads to prune tokens based on confidence.
We compare against~\citet{tang2023dynamic} in our experiments.
\citet{liu2024revisiting}~use 2-layer MLPs for token pruning in object detection and instance segmentation, achieving moderate speedups of up to $34$\% in small networks. Instead, we omit extra layers at inference time and apply \methodname to a larger ViT, achieving a $1.9\times$ speedup and $63.0$ AP\textsuperscript{box}.
We believe that \methodname significantly advances the state of the art in token pruning by being fast, maintaining high performance, and being applicable to various vision tasks.

\vspace{-0.51em}
\paragraph{Token merging.}
The assumption behind token merging is that similar token representations contribute redundantly and can thus be combined. 
Hard merging methods combine similar tokens into non-overlapping groups, e.g. through clustering~\cite{marin2023token, du2016study, zeng2022not} or bipartite matching~\cite{bolyatoken}.
In contrast, soft merging methods create summary tokens by learning convex combinations of spatial tokens~\cite{renggli2022learning, jaegle2022perceiver, haurum2022multi, zong2022self}.
For instance,~\citet{renggli2022learning} and \citet{jaegle2022perceiver} employ cross-attention with learnable queries for this purpose.
We also use cross-attention with learnable queries, but for token selection as opposed to merging, where attention scores reflect task relevance and aggregated tokens are used only in the auxiliary heads.

\vspace{-0.51em}
\paragraph{Pruning and merging.} 
Recently, pruning and merging concepts have also been applied jointly~\cite{kim2024token, wu2023ppt, bonnaerens2023learned}.
Many pruning methods additionally aggregate pruned tokens into one or a few new tokens~\cite{liang2022not, kong2022spvit, long2023beyond, xu2022evo, wei2023joint}.
\methodname similarly reactivates pruned tokens but by simply concatenating them with retained tokens before the final transformer block, without resorting to any token summarization.

%% file: sec/3_Methods.tex
\section{Methods}
\label{sec:method}

\begin{figure}[t]
  \centering
  \includegraphics[width=0.68\linewidth]{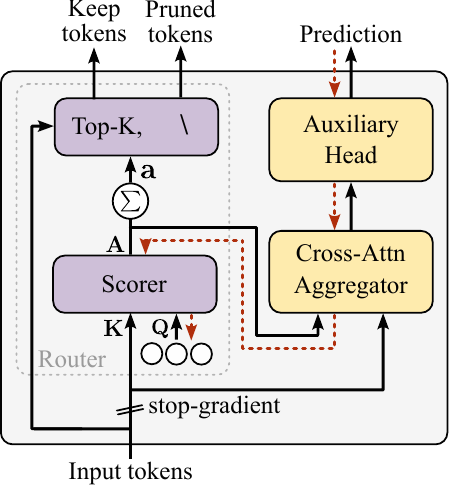}
   \caption{\methodname module during training. The router scores and separates salient keep tokens from uninformative tokens to be pruned. The scorer's attention matrix, $\mathbf{A}$, is reused in the aggregator whose output is used to make intermediate predictions. Gradient flow indicated as a dotted red line feeds back into the scorer and queries.}
   \label{fig:cropr_mod}
\end{figure}

Given a sequence of per-patch tokens, our goal is to increase the inference efficiency of ViTs by successively reducing the number of tokens as they propagate through the network.
To this end, we add \methodname modules on top of ViT blocks, each of which selects the most discriminative tokens while pruning the least informative ones.
In this way, computation in subsequent layers is reduced while relevant information is preserved, minimizing the impact of pruning on task performance.
To select the most discriminative tokens, \methodname modules use a cross-attention based routing and aggregation mechanism that receives task-specific training signals from an auxiliary head (\cref{ssec:method_cropr}). 
By slightly customizing these components, our method can be applied to various vision tasks, such as image classification, segmentation, and object detection (\cref{ssec:method_tasks}).
In particular, models for dense tasks such as semantic segmentation make pixel-wise predictions and thus require information from all tokens.
We propose Last Layer Fusion (LLF) as a simple but effective approach to recover information from pruned tokens (\cref{ssec:method_llf}).
During inference, it is possible to introduce further optimizations to slim down our module and improve throughput (\cref{ssec:method_inference}).
We end this section with a realistic example that illustrates our pruning schedule (\cref{ssec:method_example}). An implementation of \methodname is provided at: \url{https://github.com/benbergner/cropr}.

\subsection{Module description}
\label{ssec:method_cropr}

The \methodname module 
is illustrated in~\cref{fig:cropr_mod}.
Each module takes tokens $\mathbf{X}\in\R^{M\times D}$ as input and outputs disjoint sets of ``keep''  and ``pruned'' tokens, $\mathbf{X}^{k}\in\R^{K\times D}$ and $\mathbf{X}^{p}\in\R^{R\times D}$ respectively, where $K=M-R$  and $R$ is the pruning rate.
Each module consists of four components: a scorer and a selector, which together form the router, as well as an aggregator and a task head.

The \textbf{scorer} assigns scores $\mathbf{a}\in\R^{M}$ to the set of tokens.
These scores are then passed to the \textbf{selector}, which retains the $K$ highest scoring tokens and prunes the remaining $R$ tokens:
\begin{equation}
\mathbf{X}^{k}=\text{Top-K}\left(\mathbf{X} \mid \mathbf{a}\right)\label{eq:topk},
\end{equation} 
\begin{equation}
\mathbf{X}^{p}=\mathbf{X}\setminus\mathbf{X}^{k},
\end{equation}
where $\setminus$ is set subtraction.
The scorer itself is modeled after a cross-attention module with learnable queries,  $\mathbf{Q}\in\R^{N\times D}$.
The key matrix, $\mathbf{K}\in\R^{M\times D}$, is conditioned on the input tokens $\mathbf{X}$.
\begin{equation}
\mathbf{A}=\mathbf{Q} \times \mathbf{K}\!\left(\mathbf{X}\right)^{\top}. \label{eq:crs_attn}
\end{equation}
Cross-attn modules typically use linear query, key and value projections, multiple attention heads and a LayerNorm (LN)~\cite{dosovitskiy2021an, vaswani2017attention}.
We found that neither of these components is necessary for achieving high task performance in our setting. This allows us to streamline our module while increasing throughput (\cref{tab:cross-attn}).
We map $\mathbf{A}$ to $\mathbf{a}$ by summing the attention matrix over the query axis. For $N>1$,
\begin{equation}
\mathbf{a}=\sum_{n=1}^{N}\mathbf{A}_{n}, \qquad \mathbf{A}_{n}\in\R^M. \label{eq:aggr_attn_scores}
\end{equation}
This concludes the \textbf{router} design.

To learn scores that reflect a token's contribution to a prediction, the \textbf{aggregator} uses the attention matrix $\mathbf{A}$ to compute weighted averages of the input tokens, which are then passed to an auxiliary head.
Thus, over the course of training, the scorer will assign more weight to tokens that are discriminative, and these tokens will then be retained for processing by the following transformer blocks.
We found it beneficial to increase capacity in the aggregator by incorporating the transformer block's feed-forward module, adding a LN and an MLP with a residual connection (\cref{tab:mlp}).
Thus, for $\mathbf{X}^{\prime}=\text{softmax}\left(\frac{\mathbf{A}}{\sqrt{D}}\right)\mathbf{X}$,
\begin{equation}
\text{aggregator}\left(\mathbf{X} | \mathbf{A}\right)=\text{MLP}(\text{LN}(\mathbf{X}^{\prime})) + \mathbf{X}^{\prime}.
\end{equation}
Aggregated outputs are processed by task-specific heads to make intermediate predictions, which in turn provide gradients for training the aggregator and scorer.

Finally, note that a stop-gradient is applied before the scoring and aggregation blocks.
Conceptually, this has the advantage of isolating the auxiliary heads from the backbone.
Thus, the encoder is not affected by conflicting gradients from auxiliary losses.
This is also computationally efficient during training, since gradients from \methodname components do not backprop through the encoder.

\subsection{Task-specific designs}
\label{ssec:method_tasks}

Our scorer and aggregator employ a flexible query mechanism, similar to that of Perceiver IO~\cite{jaegle2022perceiver}, enabling arbitrary output shapes and easy adaptation to various tasks.
In this section, by adjusting the number of learnable queries, designing auxiliary heads and loss functions, we instantiate \methodname for each vision task, as follows.

\vspace{-0.4em}
\paragraph{Image classification.}
The scorer uses a single learnable query, $N=1$.
The aggregator then outputs a single token, which is processed using a LN and linear projection exactly as in the final classification head.
The latter outputs logits for all classes.
A softmax cross-entropy loss is used.

\vspace{-0.4em}
\paragraph{Semantic segmentation.}
Both main and auxiliary heads adopt the linear head of Segmenter~\cite{Strudel_2021_ICCV}.
The scorer uses one learnable query per patch token, $N=h\times w$, to obtain grid-structured representations from $\mathbf{X}$.
The aggregator output is processed using a LN and linear projection, like in image classification, but independently per-patch location, followed by a per-patch softmax cross-entropy loss.
To reduce computational complexity in the auxiliary heads, instead of upsampling the logits to the input resolution as in Segmenter, the labels are downsampled to the feature map resolution.
The downsampled labels can then be reused across \methodname modules
\vspace{-0.4em}
\paragraph{Joint detection and instance segmentation.}
We apply \methodname to Cascade Mask R-CNN~\cite{he2017mask,cai2019cascade} for this task.
But because this multi-stage detector is computationally expensive, it is less practical for use in auxiliary heads.
We propose a proxy auxiliary head and loss that provides a strong signal for both tasks: multi-label classification.
The intuition here is that object detection and instance segmentation both require identifying all object categories present in an image.
In more detail, ground-truth labels are encoded as binary vectors, where each dimension corresponds to a class's presence.
The scorer then uses a single learnable query like in image classification, $N=1$.
A LN and linear projection then map the aggregated token into as many logits as there are classes in the dataset.
A sigmoid activation function and a binary cross-entropy loss are used in this multi-label setting.

\subsection{Last Layer Fusion (LLF)}
\label{ssec:method_llf}

In dense tasks, such as semantic segmentation, predictions are made at the pixel level.
However, this is hard when a significant portion of the input is dropped.
In addition, many task heads require a spatial feature map for upsampling, which is not maintained during pruning.

We address these challenges with LLF, an efficient and effective approach that reactivates pruned tokens and preserves information from all image patches.
Specifically, pruned tokens from all \methodname modules are inserted alongside retained tokens output by the penultimate ViT block (\cref{fig:cropr_overview}) at their respective spatial locations.
In other words the pruned tokens are skipped to the final ViT block and not entirely discarded.
The final ViT block processes this combined sequence, allowing previously pruned tokens to attend to deep features of retained tokens.
We present t-SNE~\cite{vandermaaten08visualizing} plots to visualize its effect in~\cref{append:tsne}.
We disable DropPath~\cite{conf/iclr/LarssonMS17} in the final ViT block to ensure token fusion.

LLF introduces no additional parameters while outperforming other fusion methods (\cref{tab:fusion}).
Note that LLF is not specific to \methodname; in fact,
we equip several baselines with it in our experiments.

\subsection{Efficient inference}
\label{ssec:method_inference}

\begin{figure}
  \centering
  \begin{subfigure}{0.3\linewidth}
    \includegraphics[width=1.\linewidth]{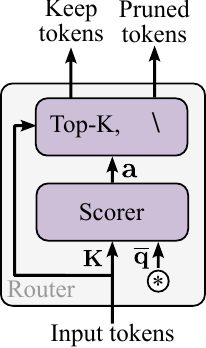}
   \caption{}
   \label{fig:cropr_inf}
  \end{subfigure}
  \hfill
  \begin{subfigure}{0.62\linewidth}
    \includegraphics[width=1.\linewidth]{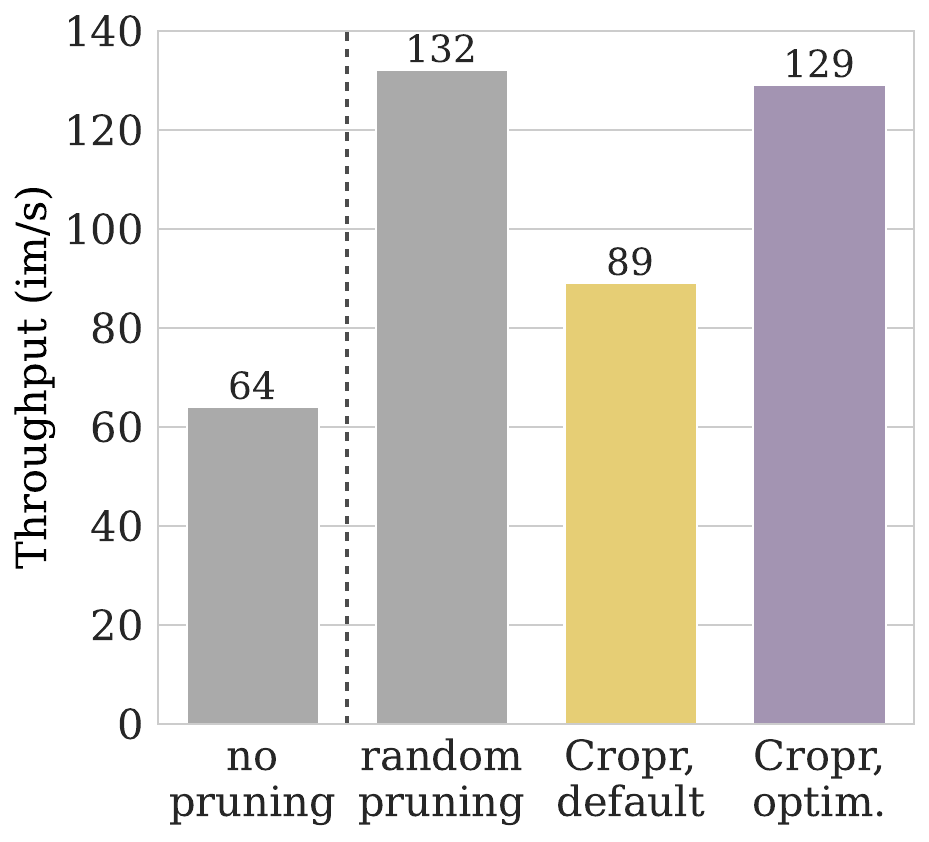}
    \caption{}
    \label{fig:cropr_inf_b}
  \end{subfigure}
  \caption{\methodname module during inference. (a) The aggregation function and the auxiliary head are removed. All queries are aggregated into a single query. (b) These optimizations speed up \methodname, with throughput comparable to that of a random selector. Results are shown for semantic segmentation.}
  \label{fig:short}
\end{figure}

The \methodname cross-attention transformer blocks and auxiliary heads constitute a significant computational overhead.
Note, however, that these components shown in yellow in~\cref{fig:cropr_mod} are only required to train the scorer.
At inference time, they can be safely discarded leaving just the router for token selection as illustrated in~\cref{fig:cropr_inf}.

The scorer still scales as $\mathcal{O}\left(N\times M\right)$, which is costly when the number of queries $N$ is large, as is the case in semantic segmentation where $N$ scales with image resolution.
But since the aggregator has now been discarded, the cross-attention matrix need not be materialized.
Only the summed up scores, $\mathbf{a}$, are needed.
Applying the distributive property of matrix multiplication, it is then easy to show that \cref{eq:aggr_attn_scores} can be reduced to a vector-matrix multiplication, $\mathcal{O}\left(M\right)$:
\begin{align}
    \mathbf{a} = \sum_{n=1}^N\left(\mathbf{Q}\mathbf{K}^\top\right)_n  &=\sum_{n=1}^N \mathbf{Q}_n \mathbf{K}^\top \\
    &= \left(\sum_{n=1}^N \mathbf{Q}_n\right)\mathbf{K}^\top =\overline{\mathbf{q}}\mathbf{K}^\top,
\end{align}
where $\overline{\mathbf{q}}\in\R^{D}$ is an aggregated query that can be precomputed.
Each \methodname module can thus be simplified to a router consisting of an efficient scoring function and a Top-K selector.
With these improvements, the throughput of \methodname is comparable to that of a random selector (\cref{fig:cropr_inf_b}).

\subsection{Pruning schedule}
\label{ssec:method_example}

We explain the pruning schedule using a working example.
Consider a ViT-L with $24$ blocks, a $224\times 224$ input image, and a patch size of $16$, resulting in $196$ patch tokens.
Unless stated otherwise, we insert \methodname modules after every block, a per-block schedule that prunes $R$ tokens at a time.
We aim to have most tokens removed by the end of the network.

Without LLF, pruning is applied after every block except the last.
In our example, setting $R=8$, we prune $23 \times 8$ 
tokens, leaving $12$ tokens, for a total pruning ratio (TPR) of $94$\%.
With LLF, pruning is performed after every block except the last two, resulting in $20$ output tokens and a TPR of $90$\%.
In this case, pruning is not performed after the pen\-ultimate block because the pruned tokens would be immediately reinserted.

We observed that for high-resolution images, maintaining the number of keep tokens as a multiple of $8$ improves throughput (\cref{append:pruning-rate}). Since patch sequence lengths are commonly divisible by $8$, we set $R$ as a multiple of $8$ whenever possible. Additionally, because ViTs typically employ a classification (CLS) token, we increase $R$ by $1$ in the first module. 
Following common practice~\cite{bolyatoken, liang2022not, fayyaz2022adaptive}, the CLS token is never pruned.

%% file: sec/4_Experiments.tex
\section{Experiments}
\label{sec:experiments}

We evaluate \methodname on four vision tasks across different ViT architectures, network capacities, and image resolutions.
To show that \methodname selects task-relevant tokens, we compare it to challenging baselines: (1) no pruning (upper bound baseline), (2) random pruning, (3) variance pruning~\cite{minderer2024scaling}, ranking tokens based on per-patch pixel variance averaged over RGB channels, and (4) Attn Top-K, which selects tokens based on self-attention scores and has been shown to be among the best performing methods~\cite{Haurum_2023_ICCV}.
For a fair comparison, we use LLF with  (2), (3), and (4).
In addition to task-specific metrics, we report FLOPs / throughput (optimal across batch sizes) for a single forward pass at inference time, using automatic mixed precision (AMP) and an NVIDIA A100 GPU.
Hyperparameters are listed in~\cref{append:hyperparam}.

\subsection{Image classification}
\label{ssec:exp_cls}
\input{tables/cls_cmp}
\paragraph{Comparison to baselines \& prior art.}

We fine-tune ViT-L on ImageNet-1k~\cite{russakovsky2015imagnet} using a pretrained masked autoencoder (MAE) following the setup of~\citet{He_2022_CVPR}, and apply the pruning schedule from our working example (\cref{ssec:method_example}). 
\Cref{tab:cls_cmp} shows a comprehensive evaluation of our method in three different scenarios.
First, we compare our method against the baselines (2) - (4).
\methodname outperforms all pruning baselines with comparable throughput.
We also include results for a non-salient selector, which inverses \methodname by pruning the most relevant tokens. As expected, this approach performs worse than random pruning. 

Next, in the middle of \cref{tab:cls_cmp}, \methodname w/o LLF is compared to prior works. 
Our method is competitive in performance and throughput.
The latter especially varies significantly across methods, with K-Medoids~\cite{marin2023token} and ATS~\cite{fayyaz2022adaptive} being slower than the unpruned baseline.

Lastly, we observe that some methods do not converge with our block-wise pruning schedule.
Hence, at the bottom of~\cref{tab:cls_cmp}, we present results for a lighter 3-stage schedule, where $R=50$ tokens are pruned after blocks $6$, $12$, and $18$, resulting in $46$ final tokens (TPR of $77$\%).
In this scenario, \methodname performs best and shows competitive throughput.

Compared to the unpruned baseline, \methodname exhibits a minor performance drop of $0.3$–$0.7$ accuracy points, while achieving a $1.6$–$1.9\times$ speedup.
In~\cref{append:pruning-rate}, we show that using a lighter schedule does not affect performance at all.

\begin{figure}
  \centering
  \includegraphics[width=1.0\linewidth]{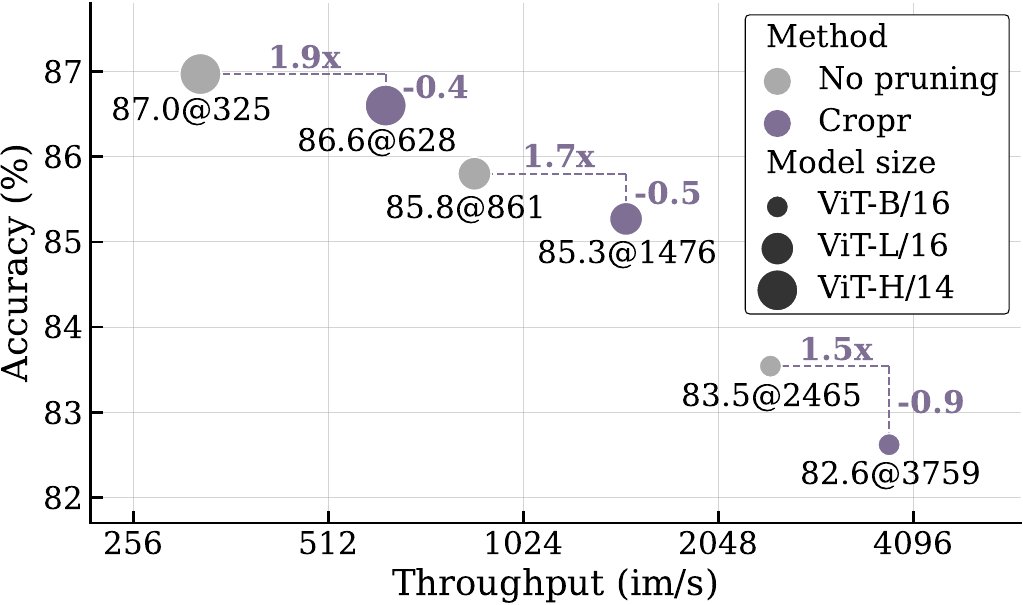}
   \caption{Performance-throughput tradeoff plot for different model sizes on ImageNet-1k. Token pruning in larger models provides more speedup and less performance drop.}
   \label{fig:model_size}
\end{figure}

\vspace{-0.1em}
\paragraph{\methodname at scale.}
What effect does network capacity have on performance and throughput in \methodname models compared to the unpruned baseline?
This question is especially relevant given the trend toward larger models~\cite{pmlr-v202-dehghani23a}.
To study this we apply \methodname with LLF to ViT-B/16, L/16, and H/14, consisting of $12$, $24$, and $32$ blocks, and set $R$ for these model sizes to $16$, $8$, and $8$ tokens per block, resulting in TPRs of $82$, $90$, and $94$\%, respectively. 
\Cref{fig:model_size} shows that the relative performance penalty of \methodname decreases as the model size increases, going from $-0.9$ in ViT-B to $-0.4$ in ViT-H, despite higher TPRs. 
This observation is likely due to the fact that in deeper models, pruning is distributed over more layers resulting in fewer tokens being dropped early on.
Furthermore, \methodname's speedup improves at scale since more layers benefit from reduced token counts, going from $1.5\times$ in ViT-B to $1.9\times$ in ViT-H.
We observe similar effects when scaling image resolution (\cref{append:img_res}).

\paragraph{Application to a SoTA model.}
We experiment with the EVA-02-L, a state-of-the-art open-source ViT~\cite{eva02}.
We start training from an IN-21K fine-tuned checkpoint, resizing images to $448\times 448$ and setting the patch size to $14$.
\Cref{tab:imagenet_sota} presents the results.
We first train EVA-02 without pruning, resulting in an accuracy of $89.9$\%, which is comparable to the $90.0$\% reported in~\citet{eva02}.
We then train a \methodname pruned version where we set $R=40$ and enable LLF, resulting in a TPR of $86$\%.
We observe an accuracy of $89.7$\%, a drop of only $0.2$ percentage points, while being $2.1\times$ faster with $\sim41\%$ fewer FLOPs.

In addition, we report results for a more aggressive pruning schedule without LLF (marked $\downarrow$), where a single \methodname module, applied after the 3rd block, prunes $825$ tokens ($80$\% of the total).
Compared to the unpruned baseline, this results in a moderate drop of $1.1$ percentage points, but provides a FLOP reduction of $\sim76$\% and a speedup of $4.1\times$.

\input{tables/imagenet_sota}

The rest of~\cref{tab:imagenet_sota} lists results for a selection of other state-of-the-art models.
After pruning, EVA-02 remains in 3rd place for accuracy, while being twice as fast.
With the more aggressive schedule, our model is the fastest by a large margin, while still outperforming some of the other models.

\subsection{Semantic segmentation}
\label{ss:ss}

We experiment on the ADE20k dataset~\cite{zhou2017scene} 
and fine-tune Segmenter~\cite{Strudel_2021_ICCV} with a linear decoding head (see~\cref{ssec:method_tasks}).
The encoder is replaced with EVA-02-L, following the settings of~\citet{eva02}.
Images are processed at a resolution of $512\times 512$ with a patch size of $16$, resulting in $1024$ patches. 
Models are trained for $64$ epochs.
For evaluation, we resize the max edge to $512$ px and pad the smaller edge while maintaining the aspect ratio.
This 1-shot evaluation approach is optimized for throughout and is more challenging than the common single-scale evaluation setting, which averages predictions from a sliding window.

In this setting the unpruned model achieves 56.7\% median mIoU across 5 seeds, outperforming Seg-L-Mask/16 ($51.8\%$ mIoU~\cite{Strudel_2021_ICCV}), despite Seg-L-Mask/16 operating at a higher resolution of $640\times640$, using a more complex mask transformer decoder, and employing the simpler single-scale evaluation setting.
We attribute this to our use of the EVA-02 pretrained backbone.

When applying \methodname, we activate LLF and prune $R=40$ tokens after each of the first $22$ blocks, resulting in a TPR of $86$\%.
To facilitate learning, a curriculum over $R$ is used for the first $32$ epochs, increasing $R$ linearly from $1$ to $40$. 

\paragraph{Comparison to baselines.}
\begin{figure}[t]
  \centering
  \includegraphics[width=1.0\linewidth]{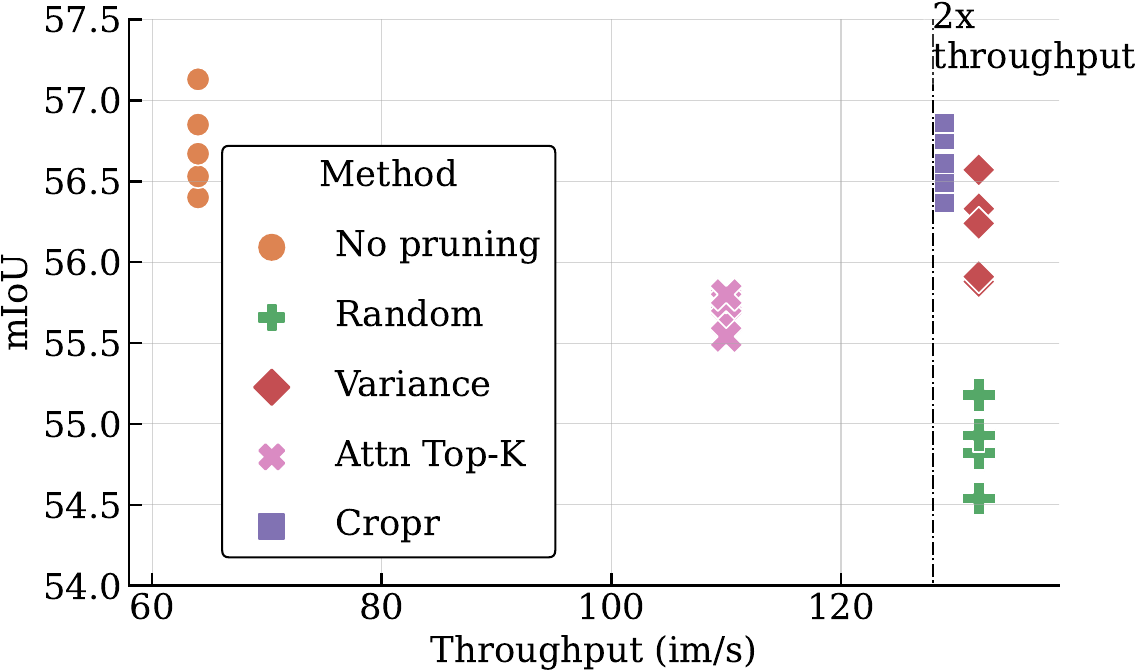}
   \caption{Semantic segmentation results on ADE20k. \methodname performs comparable to the unpruned baseline, while achieving a $2\times$ speedup, marked using the dashed vertical line. 5 seeds / method.}
   \label{fig:segm_boxplot}
\end{figure}

We compare \methodname to baselines, (1)\,--\,(4). The pruning baselines use LLF to be applicable in a segmentation setting. 
Further, Attn Top-K now uses the averaged self-attention matrix to score patches, as the CLS token is not used in the head.
Each model is run five times with different random seeds, and the  results are summarized in~\cref{fig:segm_boxplot}.
\methodname scores a median mIoU of $56.6\%$, which is only $0.1\%$ points worse than the no-pruning baseline, while being $2.0\times$ faster.
Furthermore, our model exhibits a higher median performance compared to all pruning baselines at a similar throughput.
Interestingly, we found that all baselines, even a random pruner, achieve decent performance by leveraging LLF.

\paragraph{Comparison to prior works.}
We reimplement DToP's logit fusion approach~\cite{tang2023dynamic}, but using the same settings as our method for a fair comparison.
DToP uses auxiliary heads to select tokens based on prediction confidence, and then concatenates the logits of both pruned and retained tokens to obtain the final prediction.
As in our method, we use a LN and a linear output projection as auxiliary heads.
Unlike DToP, LLF fuses features rather than logits, and the gradients from \methodname components do not backprop through the encoder. 
\Cref{tab:fusion} shows that Cropr with LLF clearly outperforms DToP's logit fusion approach.

\citet{liu2024dynamic} also use auxiliary heads, but concatenate pruned and retained token features prior to the task head.
We compare to this fusion approach in~\cref{tab:fusion}, showing that LLF outperforms `Token Concat'.
Moreover, they apply token pruning to a ViT-S-based Segmenter on ADE20k and report a $0.35$ drop in mIoU with a $18\%$ FLOP reduction relative to the no-pruning baseline.
We achieve a median drop of only $0.1$ mIoU while reducing FLOPs by $41$\%.

\begin{figure}[t]
  \centering
  \includegraphics[width=1.\linewidth]{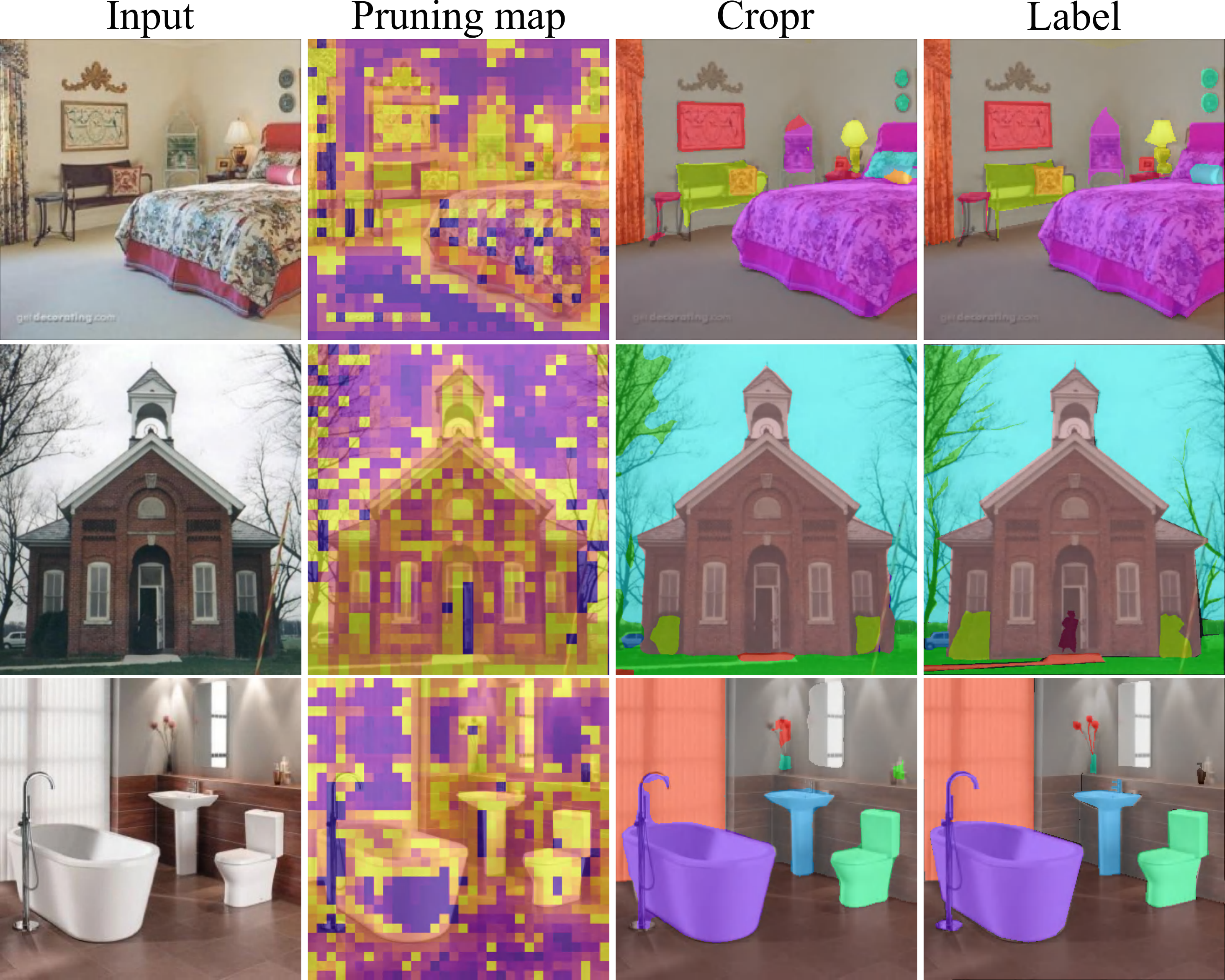}
   \caption{Visualizations for semantic segmentation. Cropr prunes tokens from stuff classes (e.g., sky, floor, wall) earlier, but keeps a few tokens from each class in later layers. Despite pruning, adjacent outputs of the same class appear consistent.}   
   \label{fig:segm_qual}
\end{figure}

\paragraph{Qualitative evaluation.}
\Cref{fig:segm_qual}~shows pruning heat\-maps for indoor and outdoor scenes.
In this task outputs for each pixel contribute to the evaluation metric, making it challenging to determine which information to prune.
We found that attention is primarily directed to salient objects; however, a few background patches are also retained in later layers, likely due to their overall relevance to the task.
Furthermore, despite pruning, we observe consistent predictions even for smaller, difficult to segment objects. This is likely facilitated by LLF, which enables early-pruned tokens to attend to deeper representations of neighboring tokens.

\subsection{Object detection and instance segmentation}
\label{ssec:det}
We benchmark \methodname on COCO~\cite{lin2014microsoft} using the EVA-02-L backbone, initialized from an Objects365~\cite{shao2019objects365} fine-tuned checkpoint.
Following~\citet{eva02}, we use Cascade Mask R-CNN as the task head to support both detection and segmentation.
Images are resized to $1536\times 1536$ with patch size $16$, yielding $96^2=9216$ patches.
As in~\citet{eva02}, global attention is intermixed with window attention. 
With a window size of $16$, this yields an initial grid of $6\times6$ windows.
To support pruning while maintaining the window size, a 5-stage pruning schedule is applied.
At each stage $i$, the number of tokens is reduced to $(96-i*16)^2$, resulting in a TPR of $97\%$.
Pruning occurs after blocks $5$, $8$, $11$, $14$, and $20$, just before the global attention layers.
LLF is applied, allowing the task head to be used without modifications.

\Cref{tab:det_cmp} shows that \methodname outperforms baselines (2)\,--\,(4) in both detection and segmentation.
The performance gap between \methodname and the unpruned model is moderate, which is expected given the high TPR.
Despite the optimized window-attention-based architecture, \methodname achieves a $54$\% reduction in FLOPs (Unpruned baseline: 2790 GFlops vs. \methodname: 1273 GFlops), along with a $2.4\times$ speedup in the encoder and a $1.9\times$ speedup in the overall model.

\begin{figure}[t]
  \centering
  \includegraphics[width=1.0\linewidth]{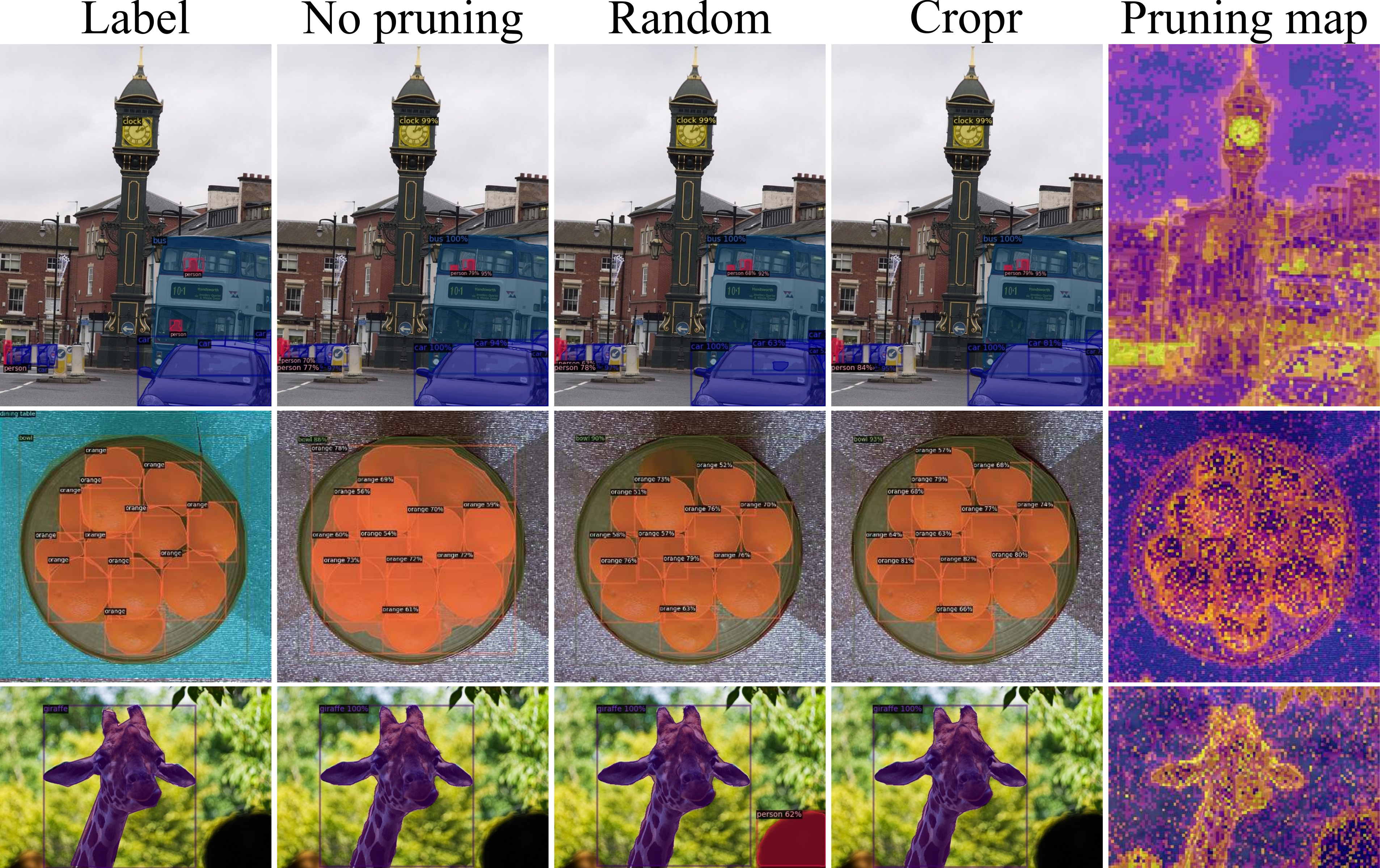}
   \caption{Bounding box and instance segmentation predictions for \methodname, as well as the unpruned and random baselines. \methodname pruning maps highlight relevant objects. First row: All methods accurately detect most objects. Second row: Only \methodname detects all oranges. Third row: Random pruner incorrectly detects a person.}
   \label{fig:det_viz}
\end{figure}

\Cref{fig:det_viz} demonstrates that \methodname modules focus on task-relevant image regions corresponding to target objects.
Interestingly, even the random pruner can serve as an effective detector with LLF, albeit with more errors. 

\input{tables/det_cmp}

\subsection{Ablation study}
\label{ssec:ablation}
\paragraph{\methodname module design.}
In~\cref{tab:cross-attn}, we compare our simplified cross-attention design (single head, w/o QKV and head projections, w/o LN) to a more complex MHA design (16 heads, w/ QKV and head projections, w/ LN).
The simpler approach outperforms MHA in both efficiency and performance metrics.
In~\cref{tab:sel-meth}, we evaluate an alternative selection method, which samples without replacement from the cross-attention distribution.
Sampling is less effective than Top-K.
\Cref{tab:mlp} indicates that incorporating an MLP (w/ LN and residual connection) into the aggregation module improves token selection. 
Crucially, this modification does not impact efficiency metrics at inference time, as the aggregator is removed after training.
Finally,~\cref{tab:grad_mode} shows that stopping the gradient flow in \methodname leads to improved results, likely because gradient interference is prevented.

\paragraph{Token fusion.}
We compare LLF to several alternatives in~\cref{tab:fusion}:
`Cross-Attn' applies a cross-attention block with grid-shaped learned queries cross-attending into the keep tokens output by the last layer, that is, pruned tokens are not reactivated.
Further note that this cross-attention block is trained from scratch.
`Token Concat' reactivates pruned tokens by concatenating them after the last layer.
`Cross-Attn + Concat' combines the two, cross-attending into concatenated tokens after the last layer.
`MHSA + Concat' is similar but uses a full self-attention transformer block  trained from scratch instead.
Lastly, `DToP' is the logit fusion approach discussed in~\cref{ss:ss}.
All methods are evaluated for semantic segmentation on the ADE20k dataset.

Not reactivating pruned tokens, as in `Cross-Attn' performs very poorly.
`Token Concat', `Cross-Attn + Concat' and `DToP' reactivate pruned tokens but do not support self-attention between pruned and retained tokens and thus underperform.
In contrast `MHSA + Concat' and LLF allow attention between tokens, resulting in higher mIoU.
Notably, LLF outperforms MHSA without introducing any additional parameters compared to the unpruned baseline.

\input{tables/design_ablations}
\input{tables/fusion}

%% file: tables/cls_cmp.tex
\begin{table}[t]
\begin{small}
\centering
\begin{tabular}{ll@{\hspace{1em}}lllr}
         Method & Sch. & LLF & Pool & Acc. & \multicolumn{1}{c}{1000 im/s}\\
     \toprule
         \textcolor[HTML]{999999}{No pruning} &  \textcolor[HTML]{999999}{---} & \textcolor[HTML]{999999}{---} & \textcolor[HTML]{999999}{avg} & \textcolor[HTML]{999999}{$85.8$} & \textcolor[HTML]{999999}{\ebar{0.86}{1.0}} \\
     \midrule
         Non-salient & \drarrow & \checkmark & avg & $76.4$ & \ebar{1.48}{1.7} \\
         Random & \drarrow & \checkmark & avg & $83.8$ & \ebar{1.50}{1.7} \\
         Variance~\cite{minderer2024scaling} & \drarrow & \checkmark & avg & $84.3$ & \ebar{1.50}{1.7} \\
         Attn Top-K & \drarrow & \checkmark & cls & $84.7$ & \ebar{1.45}{1.7} \\
         \rowcolor{bg-blue}
         \rule{0pt}{2.3ex}\textbf{\methodname} & \drarrow & \checkmark & avg & $\boldsymbol{85.3}$ & \ebar{1.48}{1.7} \\
         \aboverulesepcolor{bg-blue}
     \midrule
         K-Medoids~\cite{marin2023token} & \drarrow &  & avg & $84.5$ & \ebar{0.31}{0.4} \\
         ATS~\cite{fayyaz2022adaptive} & \drarrow &  & cls & $83.9$ & \ebar{0.49}{0.6} \\
         DPC-KNN~\cite{du2016study} & \drarrow &  & avg & $79.2$ & \ebar{1.00}{1.2} \\
         EViT~\cite{liang2022not} & \drarrow &  & cls & $84.5$ & \ebar{1.57}{1.8} \\ 
         ToMe, from~\cite{bolyatoken} & \drarrow &  & cls & $\boldsymbol{85.1}$ & \ebar{1.55}{1.8}\\ 
         ToMe~\cite{bolyatoken} & \drarrow &  & avg & $85.0$ & \ebar{1.55}{1.8} \\
     \rowcolor{bg-blue}
         \rule{0pt}{2.3ex}\textbf{\methodname} & \drarrow &  & avg & $\boldsymbol{85.1}$ & \ebar{1.61}{1.9} \\
     \aboverulesepcolor{bg-blue}
     \midrule
     DynamicViT~\cite{rao2021dynamicvit} & \stair &  & avg & $64.4$ & \ebar{1.32}{1.5} \\
         SiT~\cite{zong2022self} & \stair &  & avg & $83.0$ & \ebar{1.41}{1.6}\\ 
         Sinkhorn~\cite{haurum2022multi} & \stair &  & avg & $56.5$ & \ebar{1.40}{1.6} \\ 
         PatchMerger~\cite{renggli2022learning} & \stair &  & avg & $82.4$ & \ebar{1.40}{1.6}\\ 
     \rowcolor{bg-blue}
         \rule{0pt}{2.5ex}\textbf{\methodname} & \stair &  & avg & $85.4$ & \ebar{1.43}{1.7} \\
     \rowcolor{bg-blue}
         \textbf{\methodname} & \stair & \checkmark & avg & $\boldsymbol{85.5}$ & \ebar{1.35}{1.6} \\
     \aboverulesepcolor{bg-blue}
   \end{tabular}
 \end{small}
   \caption{ImageNet-1k results. Following~\citet{He_2022_CVPR}, we use average pooling, only reverting to CLS pooling if a method requires the CLS token. \methodname is competitive or outperforms other pruning and merging methods while being runtime-efficient. \drarrow\ : $R=8$. \stair\ : $R=50$, prune after $\{6,12,18\}$-th block.}
   \label{tab:cls_cmp}
 \end{table}

%% file: tables/imagenet_sota.tex
\begin{table}[t]
\centering
    \begin{small}
    \begin{tabular}{llll@{\hspace{1.em}}lll}
        Method & Res & \#Par & FLOPs & Acc. & \multicolumn{1}{l}{im/s}\\
    \toprule
        \textcolor[HTML]{999999}{EVA-02} & \textcolor[HTML]{999999}{$448$} & \textcolor[HTML]{999999}{$0.3$B} & \textcolor[HTML]{999999}{$0.31$B} & \textcolor[HTML]{999999}{$89.9$} & \textcolor[HTML]{999999}{$64$} \\
        \rowcolor{bg-blue}
        \rule{0pt}{2.5ex}EVA-02 + Cropr & $448$ & $0.3$B & $0.18$B & $89.7$ & $132$\\
        \rowcolor{bg-blue}
        \rule{0pt}{2.5ex}EVA-02 + Cropr $\downarrow$ 
        & $448$ & $0.3$B & $0.07$B & $88.8$ & $259$ \\
        \aboverulesepcolor{bg-blue}
    \midrule
        CAFormer-B36~\cite{yu2023metaformer} & $384$ & $0.1$B & $0.07$B & $88.1$ & $187$ \\
        RegNetY 128GF~\cite{Radosavovic_2020_CVPR} & $384$ & $0.6$B &  $0.38$B & $88.2$ & $148$ \\
        EfficientNet-L2~\cite{Xie_2020_CVPR} & $800$ & $0.5$B & $0.48$B & $88.4$ & $33$ \\
        ConvNeXt V2-H~\cite{woo2023convnext} & $512$ & $0.7$B & $0.60$B & $88.9$ & $60$ \\
        BEiT v2 ViT-L~\cite{peng2022beit}& $384$ & $0.3$B & $0.18$B & $89.0$ & $193$ \\
        MaxViT-XL~\cite{tu2022maxvit} & $512$ & $0.5$B & $0.54$B & $89.5$ & $44$\\
        ViT-L, distilled~\cite{dehghani2023scaling} & $384$ & $0.3$B & $0.18$B & $89.6$ & $193$ \\
        BEiT-3 ViT-g/14~\cite{wang2023image} & $336$ & $1.0$B & $0.58$B & $89.6$ & $85$ \\
        DaViT-Giant~\cite{ding2022davit} & $512$ & $1.4$B & $1.04$B & $90.4$ & $52$ \\
        ViT-G/14~\cite{zhai2022scaling} & $518$ & $1.8$B & $2.52$B & $90.5$ & $20$ \\ 
  \end{tabular}
  \end{small}
  \caption{Comparison of ImageNet-1k classification models. Our \colorbox{bg-blue}{EVA-02 + Cropr} variants remain competitive with SoTA models and achieve speedups of $2-4\times$ with small performance drops compared to the upper-bound baseline, EVA-02. $\downarrow$ : prune $80$\% of all tokens after the 3rd block, w/o LLF.}
  \label{tab:imagenet_sota}
\end{table}

%% file: tables/det_cmp.tex
\begin{table}[t]
\centering
\begin{small}
\begin{tabular}{lllrr}
        Method & AP\textsuperscript{box} & AP\textsuperscript{mask} & \multicolumn{1}{c}{im/s (enc.)} & \multicolumn{1}{c}{im/s}\\
    \toprule
        \textcolor[HTML]{999999}{No pruning} & \textcolor[HTML]{999999}{$64.2$} & \textcolor[HTML]{999999}{$55.4$} & \textcolor[HTML]{999999}{\ebar{5.8}{1.0}} & \textcolor[HTML]{999999}{\ebar{4.5}{1.0}} \\
    \midrule
        Random & $60.6$ & $51.9$ & \ebar{14.0}{2.4} & \ebar{8.5}{1.9}\\
        Variance & $62.0$ & $53.0$ & \ebar{13.9}{2.4} & \ebar{8.5}{1.9}\\
        Attn Top-K & $62.6$ & $53.6$ & \ebar{10.8}{1.9} & \ebar{7.3}{1.6} \\
    \midrule
    \belowrulesepcolor{bg-blue}
    \rowcolor{bg-blue}
        \textbf{\methodname} & $\boldsymbol{63.0}$ & $\boldsymbol{54.0}$ & \ebar{13.9}{2.4} & \ebar{8.5}{1.9} \\
    \aboverulesepcolor{bg-blue}  
  \end{tabular}
\end{small}
  \caption{Object detection and instance segmentation results on COCO val, showing throughput of the encoder and overall model.}
  \label{tab:det_cmp}
\end{table}

%% file: tables/design_ablations.tex
\begin{table}[t]
\begin{small}
  \centering
  \begin{subtable}{0.61\linewidth}
    \centering
    \begin{tabular}{llll}
        Method & Acc. & GFlops & im/s\\
    \toprule
        MHA & $85.2$ & $36.8$ & $1352$ \\
        \rowcolor{bg-blue}
        Simple & $85.3$ & $34.2$ & $1476$ \\
    \end{tabular}
    \caption{\textbf{Cross-attn.} A simple 1-head cross-attention design w/o projection layers performs slightly better and is more efficient.}
    \label{tab:cross-attn}
  \end{subtable}
  \hfill
  \begin{subtable}{0.34\linewidth}
    \centering
    \begin{tabular}{ll}
        Method & Acc. \\
    \toprule
        Sampling & $85.1$ \\
        \rowcolor{bg-blue}
        Top-K & $85.3$ \\
    \end{tabular}
    \caption{\textbf{Selection methods.} Top-K vs. sampling from the attention distribution.}
    \label{tab:sel-meth}
  \end{subtable}
  \begin{subtable}{0.61\linewidth}
  \vspace{0.2cm}
    \centering
    \begin{tabular}{clll}
        MLP & Acc. & GFlops & im/s\\
    \toprule
        \xmark & $85.0$ & $34.2$ & $1476$ \\
        \rowcolor{bg-blue}
        \cmark & $85.3$ & $34.2$ & $1476$ \\
    \end{tabular}
    \caption{\textbf{MLP.} Adding MLPs to the aggregator improves performance w/o overhead at inference time.}
    \label{tab:mlp}
  \end{subtable}
  \hfill
  \begin{subtable}{0.34\linewidth}
    \vspace{0.2cm}
    \centering
    \begin{tabular}{cl}
        Stop grad. & Acc. \\
    \toprule
        \xmark & $85.0$ \\
        \rowcolor{bg-blue}
        \cmark & $85.3$ \\
    \end{tabular}
    \caption{\textbf{Gradient mode.} Stopping gradient flow works best.}
    \label{tab:grad_mode}
  \end{subtable}
  \caption{Cropr ablations on ImageNet-1k, with LLF enabled.}
  \label{fig:tables}
\end{small}
\end{table}

%% file: tables/fusion.tex
\begin{table}[t!]
\centering
\begin{small}
\begin{tabular}{llll}
        Method & \#Params & GFlops & mIoU\\
    \toprule
       \textcolor[HTML]{999999}{No pruning} & \textcolor[HTML]{999999}{$304$M} & \textcolor[HTML]{999999}{$311$} & \textcolor[HTML]{999999}{$56.7$} \\
    \midrule
        Cross-Attn & $319$M & $184$ & $49.3$ \\
        Token Concat & $304$M & $172$ & $51.8$ \\
        Cross-Attn + Concat & $319$M & $186$ & $51.1$ \\
        MHSA + Concat & $318$M & $186$ & $\underline{55.2}$ \\
        DToP & $308$M & $174$ & $50.1$ \\
        \rowcolor{bg-blue}
    \midrule
    \belowrulesepcolor{bg-blue}
        \textbf{LLF} & $304$M & $183$ & $\mathbf{56.6}$\\
    \aboverulesepcolor{bg-blue}
  \end{tabular}
\end{small}
  \caption{Token fusion ablation on ADE20k. Median mIoU across 5 seeds. LLF performs best, without additional parameters.}
  \label{tab:fusion}
\end{table}

%% file: sec/5_Conclusion.tex
\section{Conclusion}
\label{sec:conc}

The experiments show that ViTs can be accelerated with small performance penalties by pruning the least informative tokens for a given task.
We showcase the versatility of our approach by applying it beyond classification to semantic and instance segmentation, as well as object detection.
That said, it is not without limitations. We discuss these in~\cref{append:limit} within the supplementary material.

Future work could extend \methodname to additional vision tasks by adapting the auxiliary heads.
Furthermore, the token-based nature of our method suggests broader applicability to other modalities, such as language and audio.

Overall, this work makes token pruning practical through a simple yet flexible method design.
Beyond pruning, we hope to inspire further exploration of efficient attention mechanisms that target task-relevant information.

%% file: sec/X_suppl.tex
\appendix
\section{Broader Impact}
\label{append:impact}

Our method significantly increases the throughput of ViTs, making it well suited for applications that require real-time inference, such as autonomous driving, robotics, and computer-assisted medical interventions.
Our approach could also be used to accelerate high-capacity models, potentially enabling new applications that require both high performance and low latency.
Edge devices such as smartphones could benefit from decreased computation to improve battery life.
Since inference is performed repeatedly and often represents a greater cumulative cost than training, our method offers a broader potential contribution to sustainability by reducing carbon emissions.

That said, it is important to acknowledge that our method could also be misused to accelerate models for harmful applications, particularly due to the versatility of \methodname across various vision tasks.
We neither explore such applications in this paper nor intend to pursue them in future work.

Moreover, we have not evaluated our method for equi\-ta\-ble performance across demographic groups.
Just as models can have biases against certain groups, these biases can propagate to token scoring and selection.
Addressing these fairness and inclusivity concerns is critical before using token pruning methods in real-world applications.
In addition, a thorough error analysis should be conducted to iden\-tify discrepancies between the pruned and unpruned models, ensuring robust and reliable performance.

\section{Limitations}
\label{append:limit}

\paragraph{Limited hardware.} Across experiments, we report $1.5$ -- $4\times$ speedups of our method over unpruned baselines, as measured on A100 NVIDIA GPUs.
However, runtime gains may vary on other hardware accelerators.
We use \texttt{gather} operations for token selection and concatenation, whose performance is hardware dependent.

\vspace{-0.5em}
\paragraph{Gap to the no-pruning baseline.} While \methodname significantly reduces computation, it does not fully close the performance gap with unpruned baselines.
This is particularly noticeable in smaller ViTs, schedules with high TPRs
, and low-resolution images (\cref{append:img_res}).

\vspace{-0.5em}
\paragraph{Pruning schedule design.} The manuscript, and this supplementary in~\cref{append:pruning-rate}, explore a variety of pruning schedules, which required manual design and task- and model-specific adaptations.
In contrast, automated schedules, conditioned on user-defined constraints like target performance and throughput, would likely be more user-friendly.

\vspace{-0.5em}
\paragraph{Quite a few tasks but not all.} We have evaluated \methodname solely on vision tasks. As discussed in the main text, \methodname could be extended to other modalities.
Furthermore, as the title suggests we address quite a few tasks, but not all of them.
While tasks such as fine-grained recognition are a trivial application of \methodname, other tasks such as visual question answering and image retrieval require follow-up work.

\section{Hyperparameters}
\label{append:hyperparam}

In ~\cref{tab:in_mae_hyperp,tab:in_eva_hyperp,tab:ade_hyperp,tab:coco_hyperp}, we list hyperparameters for the datasets and models we use in our experiments.
These settings are adopted from~\citet{He_2022_CVPR,eva02,Strudel_2021_ICCV}.
Hyperparameter and design choices specific to \methodname are described in the main text. 

\section{Different image resolutions}
\label{append:img_res}

We investigate the effect of image size on the performance and throughput of \methodname models. 
We apply \methodname with LLF to an MAE-pretrained ViT-L on ImageNet-1k at resolutions of $224$, $336$, and $448$ pixels per side.
The pruning rate $R$ scales with image size to $8$, $18$, and $32$ tokens per block, respectively, maintaining a TPR of $90$\% across all settings.

\begin{figure}[b]
  \centering
  \includegraphics[width=1.0\linewidth]{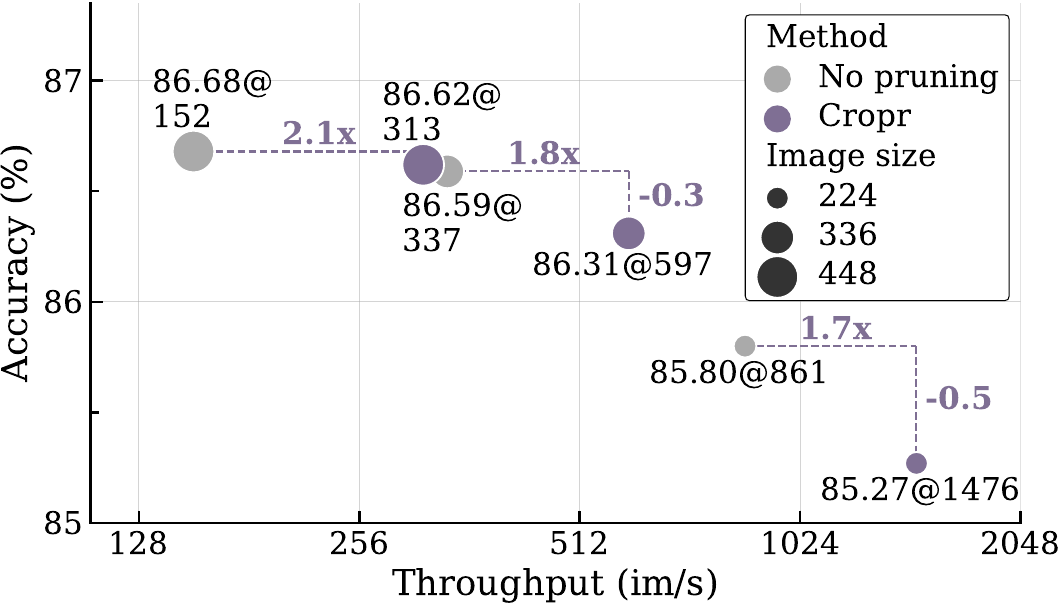}
   \caption{Performance-throughput trade-off plot for different image sizes on ImageNet-1K. Token pruning in higher-resolution images provides more speedup and less performance drop.}
   \label{fig:img_res}
\end{figure}

\Cref{fig:img_res}~shows that \methodname's relative performance penalty decreases at higher resolutions, improving from $-0.5$ to $-0.06$, effectively closing the gap to the unpruned model.
Furthermore, throughput gains are elevated at higher resolutions, going from a speedup of $1.7\times$ at $224^2$ px to a speedup of $2.1\times$ at $448^2$ px.
This is perhaps due to the quadratic relationship between sequence length and compute in transformer models.

\input{tables/hyperparams}
\clearpage

\section{Throughput ablations}
\label{append:pruning-rate}

In this section, we evaluate different pruning rates $R$, investigate the effect of keep token sequence lengths on runtime, and compare different numerical precision modes and FlashAttention~\cite{dao2022flashattention}.
ViT-L is employed for all ablations.

\vspace{-0.5em}
\paragraph{Different pruning rates.}

We ablate the pruning rate 
$R$ in our image classification setting, fine-tuning an MAE-pretrained ViT on ImageNet-1K with \methodname and LLF.
We vary the pruning rate from $R = 0$ (no pruning) to $R=8$ (value used in the manuscript).
We report top-1 accuracy and throughput in~\cref{tab:r}.
For light schedules, with $R\leq2$, performance is maintained with up to $8$\% higher throughput.
When allowing for a drop of $0.1$ accuracy points, the model can be accelerated up to $35$\% using $R=5$.

\newcommand{\minitab}[2][l]{\begin{tabular}{#1}#2\end{tabular}}
\begin{table}[t!]
\centering
\begin{small}
\begin{tabular}{llrc}
        $R$ & Acc. & \multicolumn{1}{c}{im/s} & Comments\\
    \toprule
       \textcolor[HTML]{999999}{$0$} &  \textcolor[HTML]{999999}{$85.8$} & \textcolor[HTML]{999999}{\ebar{861}{1.00}} & \multirow{3}{*}{\minitab[c]{No performance\\drop}}\\
        $1$ & $85.8$ & \ebar{883}{1.03} & \\
        $2$ & $85.8$ & \ebar{934}{1.08} & \\
    \midrule
        $3$ & $85.7$ & \ebar{996}{1.16} & \multirow{3}{*}{\minitab[c]{0.1\% Accuracy\\drop}} \\
        $4$ & $85.7$ & \ebar{1067}{1.24} &  \\
        $5$ & $85.7$ & \ebar{1160}{1.35} & \\
    \midrule
        $6$ & $85.6$ & \ebar{1244}{1.44} & \\
        $7$ & $85.5$ & \ebar{1357}{1.58} & \\
        $8$ & $85.3$ & \ebar{1476}{1.71} & \\
  \end{tabular}
\end{small}
  \caption{Accuracy and throughput for varying pruning rates on ImageNet-1k using an MAE-pretrained ViT-L.}
  \label{tab:r}
\end{table}

\vspace{-0.5em}
\paragraph{Being divisible by 8?}

Small changes in the number of tokens has a surprisingly large impact on throughput. 
We evaluated this effect across image sizes $512$, $1024$, and $2048$, with corresponding patch sequence lengths $M=1024, 4096$, and $16384$, respectively, with a patch size of $16$ (ignoring the CLS token).
\methodname is applied without LLF.

\begin{figure}[t]
  \centering
  \includegraphics[width=0.95\linewidth]{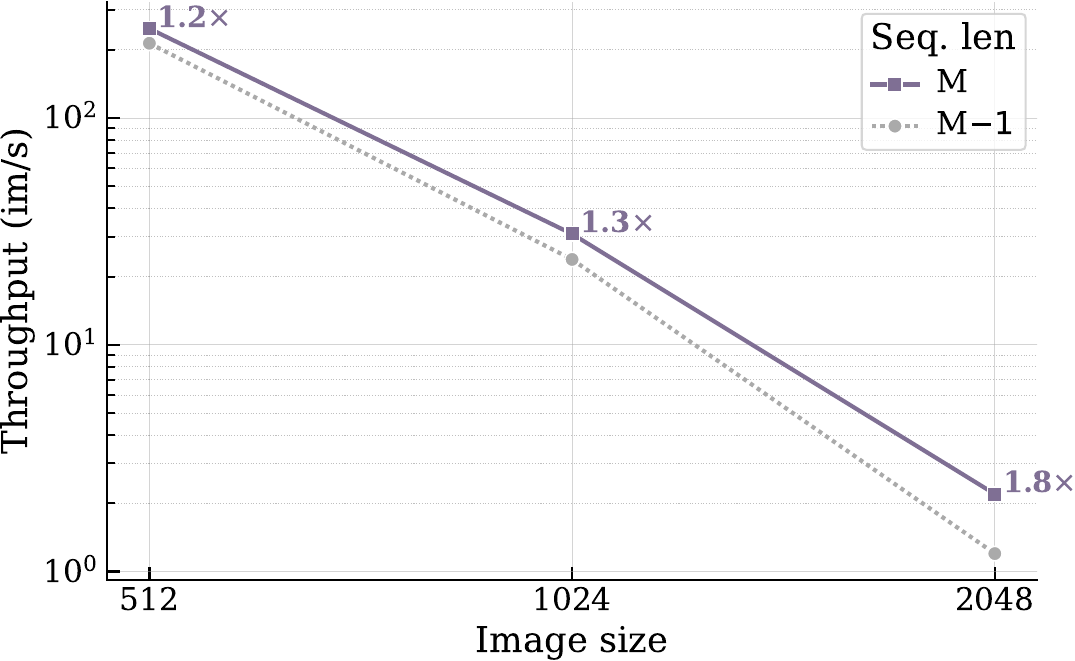}
   \caption{Effect of sequence length $M$ on throughput for different image sizes. Annotations denote speedups. A mere reduction of 1 token, instead of giving a negligible speedup, results in significant throughput drops. Both the x and y-axis are log scaled.}
   \label{fig:res_throughput_min1}
\end{figure}

We compare the throughput of two models in~\cref{fig:res_throughput_min1}.
The solid line uses pruning rates $R$ of $40$, $160$, and $640$ tokens per block for each image size respectively, resulting in a TPR of $90$\% across image sizes.
The dotted line on the other hand artificially sets the sequence lengths to $M-1$, i.e. subtracting one patch with otherwise identical settings, resulting in initial sequence lengths of $1023$, $4095$, and $16383$.

As seen in the plot, despite the reduction of one token in the dotted line case, the throughput drops significantly.
At the highest resolution, this is in fact a $1.8\times$ slowdown.
This slowdown is likely due to worse memory alignment and thread utilization in the accelerator.
We hypothesize that schedules where the number of remaining  tokens is divisible by 8 are likely to achieve the highest throughput and used that as a rule of thumb when designing pruning schedules for all our experiments.

\vspace{-0.5em}
\paragraph{Numerical precision and FlashAttention.}

In the main paper, all models were run using automatic mixed precision (AMP).
Changes to this setting primarily affect model throughput.
Here, we add to that and report throughputs for models that use (a) FP32 numerical precision, and (b) AMP in combination with FlashAttention~\cite{dao2022flashattention}.
\methodname is applied without LLF, setting $R$ as in the previous ablation to achieve a TPR of $90$\% for all image sizes.

\begin{figure}[h]
  \centering
  \includegraphics[width=1.0\linewidth]{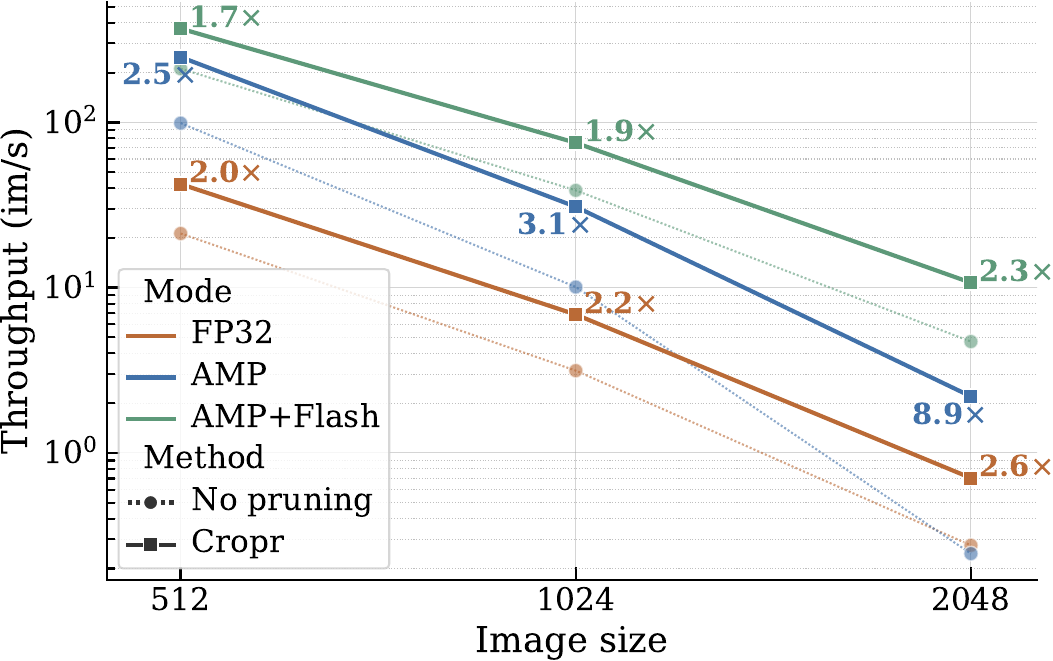}
   \caption{Throughput ablations for FP32, AMP, and AMP with FlashAttention across image sizes. Annotations denote speedups of \methodname over the unpruned baselines.}
   \label{fig:res_throughput}
\end{figure}
 
As shown in~\cref{fig:res_throughput}, 
\methodname improves over the unpruned baseline in terms of throughput in all three settings.
Relative speedups are higher for larger images, in line with the findings in~\cref{append:img_res}.
Notably, for images at a resolution of $2048^2$, \methodname achieves a speedup of up to $8.9\times$ with AMP.

AMP + Flash Attention, is the fastest setting overall.
But even in this optimized regime, \methodname delivers a significant speedup between $1.7\times$ and $2.3\times$.

\section{t-SNE visualizations of LLF's effect}
\label{append:tsne}

\begin{figure}[t]
  \centering
  \begin{subfigure}{\linewidth}
    \includegraphics[width=1.\linewidth]{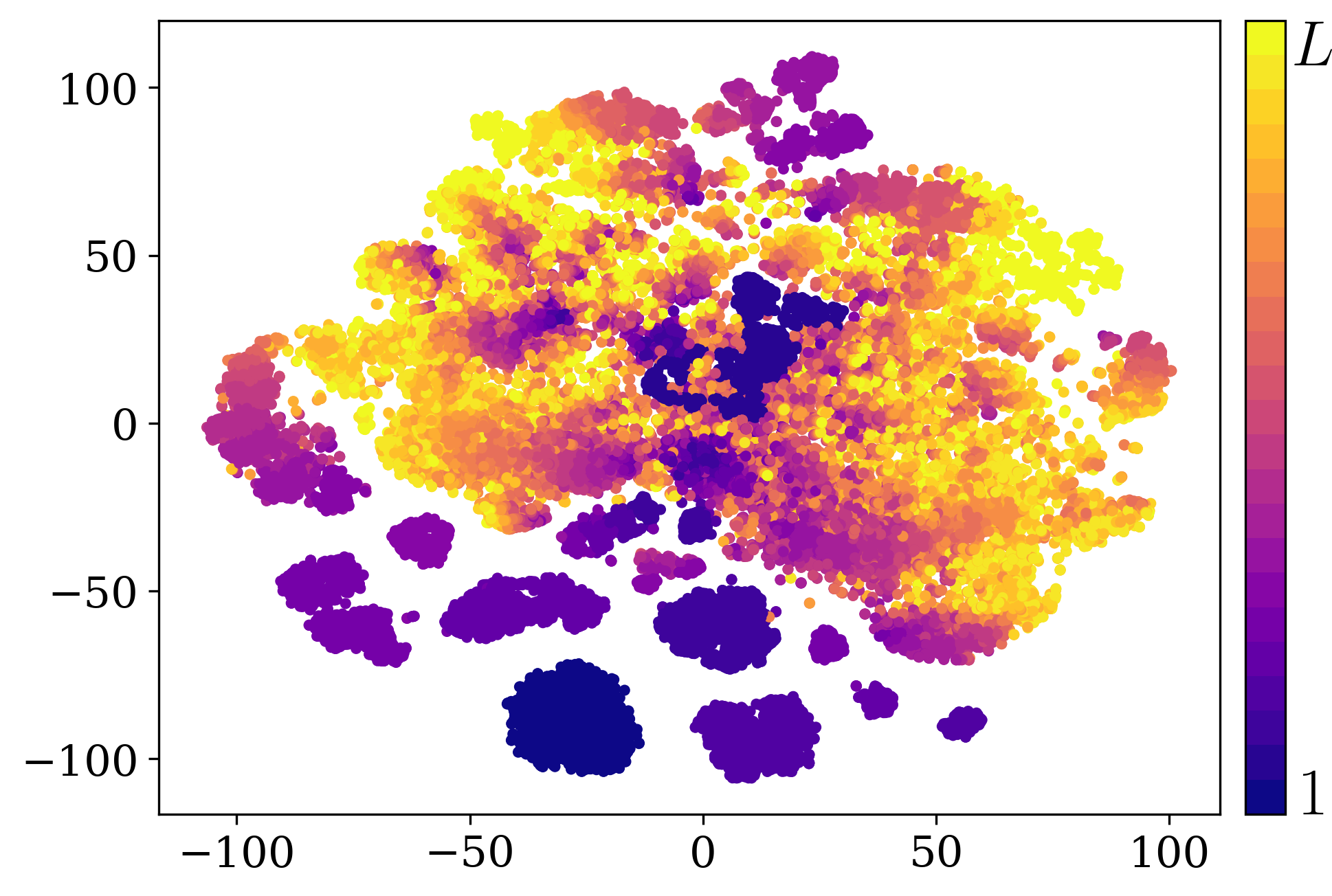}
   \caption{`Token Concat'}
   \label{fig:tsne_concat}
  \end{subfigure}
  \hfill
  \\
  \begin{subfigure}{\linewidth}
    \includegraphics[width=1.\linewidth]{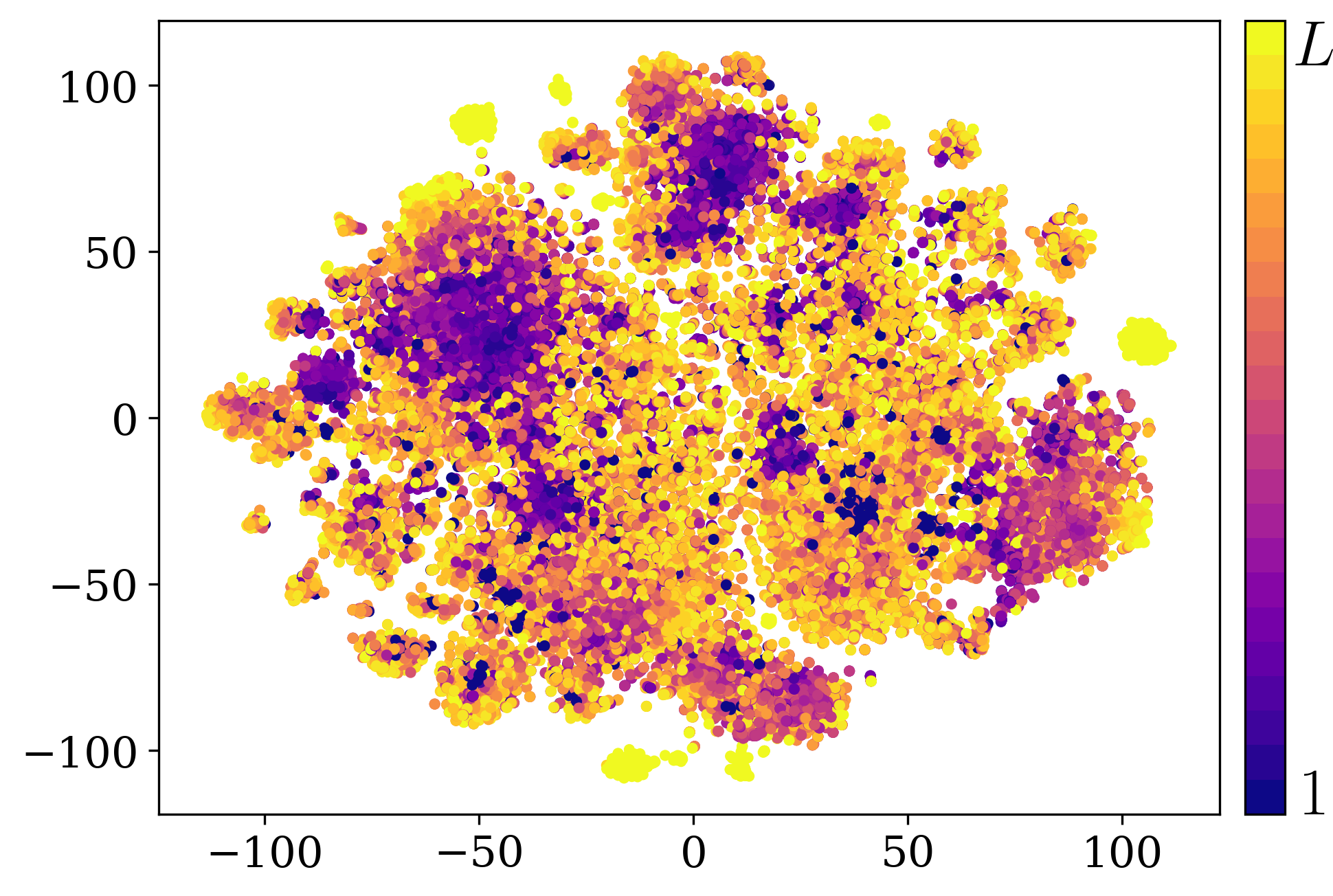}
    \caption{LLF}
    \label{fig:tsne_llf}
  \end{subfigure}
  \caption{t-SNE projections of tokens extracted right before the prediction head. Tokens are coloured according to the block after which they were pruned. We compare two fusion methods: (a) `Token Concat', (b) LLF. The latter has a more uniform distribution suggesting that LLF helped synchronize these tokens.}
  \label{fig:tsne}
\end{figure}

In ~\cref{tab:fusion}, we compared LLF and the `Token Concat' baseline.
Whereas `Token Concat' performs token concatenation after the last transformer block, LLF does it after the penultimate block, enabling the pruned tokens and kept tokens to attend into each other and to loosely speaking synchronize.
We visualize this effect in~\cref{fig:tsne} using t-SNE~\cite{van2009learning} down-projected tokens.

We apply t-SNE to the ADE20k validation set, and for visual clarity we plot only the top-1 scoring tokens within the respective pruned token sets per block.
Points are then colored according to the block number of the block after which they were pruned.
As seen in the `Token Concat' case,~\cref{fig:tsne_concat}, tokens pruned after different blocks occupy different regions in the embedding space, which might be challenging for the linear prediction head trying to map them into class labels.
In the LLF case,~\cref{fig:tsne_llf}, the embedding space is more uniformly occupied by tokens pruned at different stages, supporting our hypothesis that LLF helps synchronize these tokens.
We argue that this may be easier for the linear prediction head to then learn a projection into class logits.

%% file: tables/hyperparams.tex
\newcommand{\thickhline}{\noalign{\hrule height 0.12em}}
\begin{table}
\begin{small}
  \begin{tabular}{l|l}
        \textbf{Config} & \textbf{Value} \\
    \thickhline
        checkpoint & MAE-pretrained~\cite{He_2022_CVPR} \\
        learning rate & 4e-3 \\
        layer-wise lr decay~\cite{clark2020electra,bao2021beit} & 0.65 (B), 0.75 (L, H) \\
        learning rate schedule & cosine decay~\cite{loshchilov2017sgdr} \\
        optimizer & AdamW~\cite{loshchilov2018decoupled} \\
        optimizer hparams & $\beta_1$, $\beta_2$, $\epsilon$ = 0.9, 0.999, 1e-8 \\
        weight decay & 0.05 \\
        input size per side & 224, 336 or 448 \\
        patch size & 16 (B, L), 14 (H) \\
        batch size & 1024 \\
        epochs & 100 (B), 50 (L/H) \\
        warm-up epochs & 5 \\
        label smoothing~\cite{szegedy2016rethinking} & 0.1 \\
        drop path~\cite{huang2016deep} & 0.1 (B), 0.2 (L), 0.3 (H) \\
        augmentation & RandAug(9, 0.5)~\cite{randaugment} \\
        random resized crop & (0.08, 1) \\
        cutmix~\cite{yun2019cutmix} & 1.0 \\
        mixup~\cite{zhang2018mixup} & 0.8 \\
        CLS token & \cmark \\
   \end{tabular}
  \caption{\textbf{ImageNet-1k} image classification hyperparameters for \textbf{MAE}-pretrained encoders.}
  \label{tab:in_mae_hyperp}
\end{small}
\end{table}

\begin{table}
\begin{small}
  \begin{tabular}{l|l}
        \textbf{Config} & \textbf{Value} \\
    \thickhline
        \multirow{ 2}{*}{checkpoint}& IN-21K fine-tuned \\
         & EVA-02-L~\cite{eva02} \\
        learning rate & 2e-5 \\
        layer-wise lr decay~\cite{clark2020electra,bao2021beit} & 0.85 \\
        learning rate schedule & cosine decay~\cite{loshchilov2017sgdr} \\
        optimizer & AdamW~\cite{loshchilov2018decoupled} \\
        optimizer hparams & $\beta_1$, $\beta_2$, $\epsilon$ = 0.9, 0.999, 1e-8 \\
        weight decay & 0.05 \\
        input size per side & 448 \\
        patch size & 14 \\
        batch size & 512 \\
        epochs & 20 \\
        warm-up epochs & 2 \\
        label smoothing~\cite{szegedy2016rethinking}  & 0.2 \\
        drop path~\cite{huang2016deep} & 0.15 \\
        augmentation & RandAug(9, 0.5)~\cite{randaugment} \\
        random resized crop & (0.08, 1) \\
        cutmix~\cite{yun2019cutmix} & \xmark \\ 
        mixup~\cite{zhang2018mixup} & \xmark \\
        CLS token & \cmark \\
   \end{tabular}
  \caption{\textbf{ImageNet-1k} image classification hyperparameters for \textbf{EVA-02}-pretrained encoders.}
  \label{tab:in_eva_hyperp}
\end{small}
\end{table}

\begin{table}
\begin{small}
  \begin{tabular}{l|l}
        \textbf{Config} & \textbf{Value} \\
    \thickhline
        checkpoint & MIM pretrained EVA-02-L~\cite{eva02} \\
        learning rate & 2e-5 \\
        layer-wise lr decay~\cite{clark2020electra,bao2021beit} & 0.9 \\
        learning rate schedule & polynomial decay~\cite{deeplab} \\
        optimizer & AdamW~\cite{loshchilov2018decoupled} \\
        optimizer hparams & $\beta_1$, $\beta_2$, $\epsilon$ = 0.9, 0.999, 1e-8 \\
        weight decay & 0.05 \\
        input size per side & 512 \\
        patch size & 16 \\
        batch size & 8 \\
        epochs & 64 \\
        warm-up epochs & 0 \\
        drop path~\cite{huang2016deep} & 0.2 \\
        CLS token & \cmark \\
   \end{tabular}
  \caption{\textbf{ADE20k} semantic segmentation hyperparameters.}
  \label{tab:ade_hyperp}
\end{small}
\end{table}

\begin{table}
\begin{small}
  \begin{tabular}{l|l}
        \textbf{Config} & \textbf{Value} \\
    \thickhline
        \multirow{ 2}{*}{checkpoint}& Objects365 fine-tuned \\
         & EVA-02~\cite{eva02} \\
        learning rate & 4e-5 \\
        layer-wise lr decay~\cite{clark2020electra,bao2021beit} & 0.8 \\
        learning rate schedule & constant \\
        optimizer & AdamW~\cite{loshchilov2018decoupled} \\
        optimizer hparams & $\beta_1$, $\beta_2$, $\epsilon$ = 0.9, 0.999, 1e-8 \\
        weight decay & 0.1 \\
        input size per side & 1536 \\
        patch size & 16 \\
        batch size & 64 \\
        training steps & 40k \\
        drop path~\cite{huang2016deep} & 0.3 \\
        large-scale jittering~\cite{ghiasi2021simple} & \cmark \\
        attention window size & 16 \\
        global attn block ids & 3, 6, 9, 12, 15, 18, 21, 24 \\        
        max numbers of detection & 100 \\
        softNMS~\cite{bodla2017soft} & IoU threshold = 0.6 \\
        maskness scoring~\cite{huang2019mask,wang2021solo} & maskness threshold = 0.5 \\        
        EMA decay~\cite{polyak1992acceleration} & 0.999 \\
        CLS token & \xmark \\
   \end{tabular}
  \caption{\textbf{COCO} object detection and instance segmentation hyperparameters.}
  \label{tab:coco_hyperp}
\end{small}
\end{table}